\documentclass[conference]{IEEEtran}
\usepackage{fancyhdr}
\usepackage{graphicx}
\graphicspath{{images/}}

\usepackage{amsmath,amssymb}
\usepackage{cite}
\usepackage{booktabs}
\usepackage{multirow}
\usepackage{float}
\usepackage{makecell}
\usepackage{subcaption}
\usepackage{lettrine}
\usepackage{url}



\makeatletter
\newcommand{\blfootnote}[1]{%
  \begingroup
  \renewcommand{\thefootnote}{}%
  \footnote{#1}%
  \addtocounter{footnote}{-1}%
  \endgroup
}
\makeatother

\title{NGram-MoSE: Efficient Remote Sensing Super-Resolution via N-Gram Context and Mixture-of-Experts}

\author{
\IEEEauthorblockN{
Yun-Hsuan Huang,
Trong-An Bui,
Chih-Hung Chuang
}
\IEEEauthorblockA{\scriptsize Institute of Aerospace and Systems Engineering, National Taipei University of Technology, Taipei City, Taiwan.}
}

\begin{document}

\maketitle

\blfootnote{Corresponding author: Trong-An Bui (trongan93@ntut.edu.tw). This research is partly supported by the National Science and Technology Council (NSTC), Taiwan, under Grant Nos. NSTC 114-2221-E-027-030-MY2 and 114-2218-E-027-003.}

\begin{abstract}
Remote sensing applications for environmental monitoring and disaster management are frequently constrained by a spatial--temporal trade-off: imagery with fine spatial detail is often acquired less frequently, whereas more temporally available observations are typically coarser. Single-image super-resolution provides a practical means to enhance coarse imagery without changing acquisition schedules, yet many Transformer-based SR models remain computationally expensive and can be sensitive to limited or geographically biased training data, which degrades robustness under out-of-distribution conditions. This paper presents NGram-MoSE, a lightweight Transformer architecture designed to improve both efficiency and texture continuity. NGram-MoSE introduces N-Gram Context Injection to strengthen cross-window local consistency and mitigate window-boundary artifacts, and incorporates a Mixture-of-Experts (MoE) feed-forward design to scale capacity through sparse activation without proportional growth in inference cost. Experiments on a geographically disjoint OOD test set show that NGram-MoSE achieves 31.68\,dB PSNR while reducing FLOPs by \(14\times\) relative to a heavyweight Transformer reference. Downstream evaluation on a landslide segmentation benchmark further demonstrates that restoring degraded inputs to the detector training scale improves performance, yielding a 4.47\% absolute gain in mAP@50 over bicubic upsampling, and exhibits stronger cross-scale consistency under scale extrapolation. These results indicate that NGram-MoSE provides an effective SR module for resource-constrained remote sensing pipelines requiring robust generalization.
\end{abstract}

\section{Introduction}
\label{sec:intro}

Remote sensing (RS) is indispensable for environmental monitoring and disaster management, yet operational use is frequently constrained by a spatial--temporal trade-off. Imagery with fine spatial detail is often acquired less frequently, whereas observations with higher temporal availability are commonly delivered at coarser spatial resolution. This mismatch forces a choice between structural clarity and timely coverage, which is particularly restrictive for rapid geohazard assessment. In landslide mapping, delayed access to fine-resolution imagery can impede situational awareness, while coarse-resolution observations may obscure boundaries and local cues required for accurate delineation. Single-image super-resolution (SISR) offers a practical means to enhance coarse imagery when fine-resolution acquisitions are unavailable, improving interpretability and supporting downstream analysis without altering acquisition schedules.

The practical use of deep SISR in RS settings faces two major barriers. First, many state-of-the-art architectures are computationally expensive, limiting applicability to large-area processing and deployment in resource-constrained environments where power and memory are strictly limited. Second, high parameter density can lead to overfitting under limited or geographically biased training data, producing over-sharpening and high-frequency artifacts that reduce robustness and may degrade downstream performance. These limitations are consequential for landslide segmentation, where reliable predictions depend on coherent boundaries and physically plausible texture patterns to separate disturbed areas from complex backgrounds. Furthermore, standard window-based attention mechanisms can introduce boundary discontinuities and may fail to preserve the continuous, repetitive textures characteristic of geomorphological structures.

NGram-MoSE is presented as an efficient Transformer-based framework for RS SISR that targets both computational efficiency and semantic fidelity. The design integrates N-Gram Context Injection to model dependencies across overlapping local windows and improve texture continuity, together with a Mixture-of-Experts (MoE) feed-forward network that expands capacity through sparse expert activation. Robustness is assessed using a geographically disjoint out-of-distribution test set spanning diverse terrains, and practical utility is verified via downstream landslide segmentation on the Landslide4Sense benchmark.

The primary contributions are summarized as follows:
\begin{itemize}
    \item \textbf{Efficiency:} NGram-MoSE achieves competitive restoration quality while reducing computational cost by \(14\times\) relative to a heavyweight Transformer baseline, improving feasibility for resource-constrained RS pipelines.
    \item \textbf{Texture integrity:} N-Gram Context Injection mitigates window-boundary artifacts by explicitly modeling local cross-window dependencies, preserving terrain continuity in reconstructed outputs.
    \item \textbf{Application-driven validation:} Downstream evaluation on Landslide4Sense demonstrates improved detection performance, including a \(+4.47\%\) absolute gain in mAP@50 over bicubic upsampling under the restoration setting considered in this work.
\end{itemize}

\section{Related Work}
\label{sec:related}

Single-image super-resolution (SISR) has been widely studied in natural-image and remote sensing applications. Early deep SR methods mainly relied on CNN-based residual architectures for feature extraction and reconstruction \cite{7115171, Lim_2017_CVPR_Workshops, 7937881}. More recently, Transformer-based models such as SwinIR \cite{Liang_2021_ICCV} and HAT \cite{Chen_2023_CVPR} have improved restoration fidelity through window attention and hybrid attention designs. However, these high-capacity models often require substantial computation and may generalize poorly under limited or geographically biased remote sensing data. Moreover, non-overlapping window attention can introduce boundary discontinuities that disrupt continuous terrain textures.

Several approaches address this issue by enriching local context across neighboring regions. The N-Gram context mechanism \cite{Choi_2023_CVPR} aggregates information from overlapping patches to strengthen local continuity and reduce block-wise artifacts, which is well suited to remote sensing imagery with repetitive and spatially continuous terrain patterns.

Efficiency is another important concern for SR deployment. Lightweight models such as IMDN \cite{10.1145/3343031.3351084} reduce computation through feature distillation, but often face a trade-off between efficiency and fidelity. Mixture-of-Experts (MoE) \cite{shazeer2017outrageouslylargeneuralnetworks} offers an alternative by scaling model capacity through sparse expert activation. Recent vision MoE models, such as Swin2-MoSE \cite{Rossi_2025}, further demonstrate the potential of expert specialization for efficient Transformer-based SR.

Motivated by these studies, NGram-MoSE integrates N-Gram context aggregation with an MoE feed-forward design to improve texture continuity and efficient capacity scaling for remote sensing SR, with application-driven evaluation on landslide detection and segmentation using Landslide4Sense \cite{Ghorbanzadeh_2022}.

\section{Proposed Method}
\label{sec:method}

\begin{figure}[H]
    \centering
    \includegraphics[width=\linewidth]{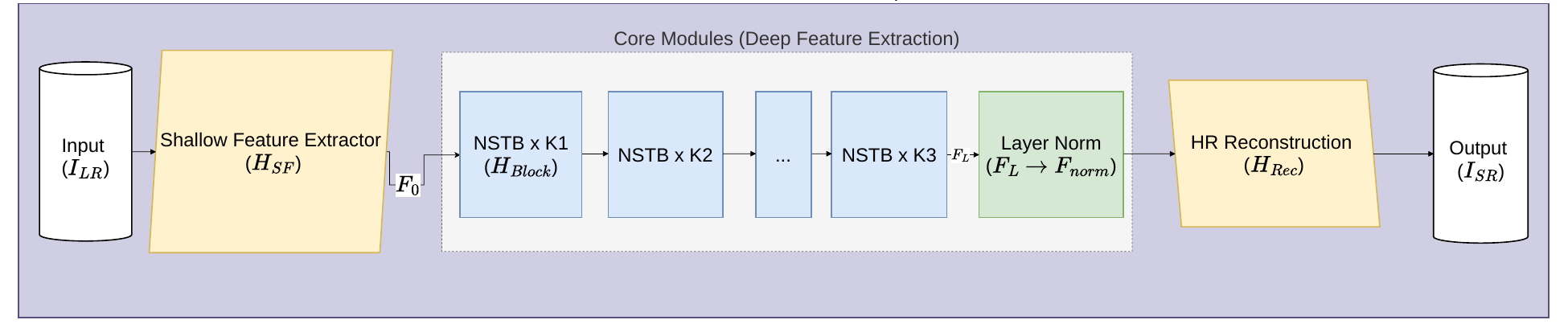}
    \caption{Overall Architecture: The proposed NGram-MoSE framework consists of a shallow feature extractor, deep feature extraction via stacked NSTBs, and a reconstruction module.}
    \label{fig:arch_overall}
\end{figure}

\begin{figure}[H]
    \centering
    \includegraphics[width=\linewidth]{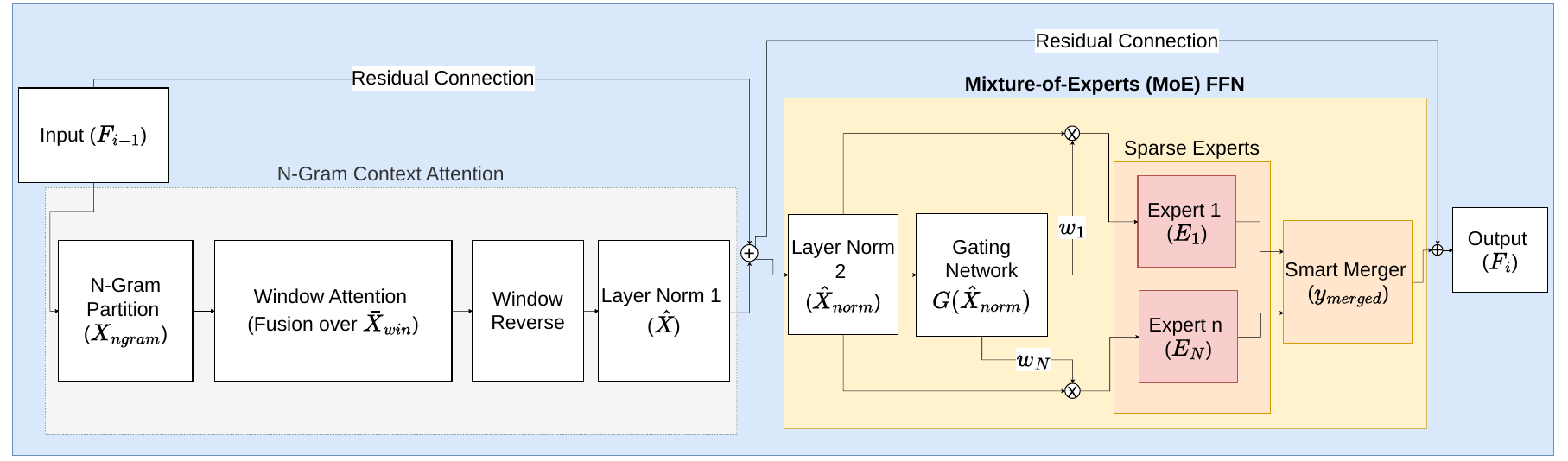}
    \caption{N-Gram Swin Transformer Block (NSTB): Integrating N-Gram Context Attention with a Mixture-of-Experts (MoE) feed-forward network.}
    \label{fig:arch_nstb}
\end{figure}

This section presents NGram-MoSE, a lightweight framework for remote sensing image super-resolution. An overview is shown in Fig.~\ref{fig:arch_overall}. The framework adopts a hierarchical pipeline with three stages: shallow feature extraction, deep feature extraction using stacked NSTB-MoSE blocks, and high-resolution reconstruction. The key building block is NSTB-MoSE (Fig.~\ref{fig:arch_nstb}), designed to address two practical limitations of window-based Transformer SR in remote sensing: (i) discontinuities and texture breaks induced by non-overlapping windows, and (ii) the high computational cost of dense feed-forward layers when model capacity is scaled. These issues are mitigated by integrating N-Gram context attention for cross-window continuity and a sparse Mixture-of-Experts (MoE) feed-forward network for conditional computation.

The shallow feature extraction stage converts the input image into an embedding space suitable for Transformer processing. A $3 \times 3$ convolution is employed to preserve local spatial structure while projecting the input into a feature representation:
\begin{equation}
  F_0 = H_{SF}(I_{LR}), \quad I_{LR} \in \mathbb{R}^{H \times W \times C_{in}},
  \label{eq:shallow_feature}
\end{equation}
where $H_{SF}(\cdot)$ denotes the shallow feature extractor. The deep feature extraction stage then increases receptive field and representation power through $L$ stacked NSTB-MoSE blocks, enabling progressive refinement of high-frequency details:
\begin{equation}
  F_i = H_{Block_i}(F_{i-1}), \quad i = 1, \dots, L .
  \label{eq:deep_feature}
\end{equation}

Before upsampling, feature normalization is applied to reduce distribution shift across blocks and stabilize reconstruction, which is particularly important when sparse expert routing is used in the feed-forward pathway:
\begin{equation}
  F_{\text{norm}} = \text{LayerNorm}(F_L).
  \label{eq:norm}
\end{equation}
A reconstruction head then produces the super-resolved output using sub-pixel upsampling, which is computationally efficient and avoids checkerboard artifacts associated with some deconvolution operators:
\begin{equation}
  I_{SR} = H_{Rec}(F_{\text{norm}}) = \text{Conv}\!\left(\text{PixelShuffle}\!\left(\text{Conv}(F_{\text{norm}})\right)\right).
  \label{eq:reconstruction}
\end{equation}

\begin{figure}[H]
    \centering
    \includegraphics[width=\linewidth]{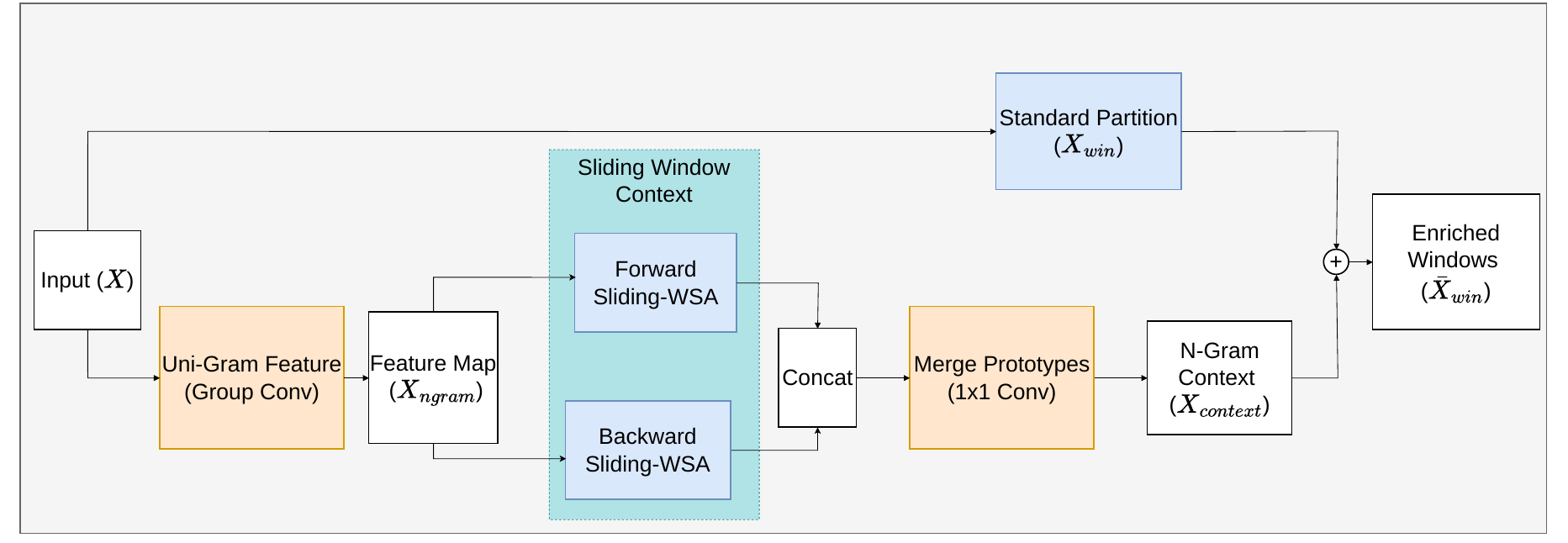}
    \caption{N-Gram Context Module: Capturing sliding-window contexts to enhance local continuity across attention windows.}
    \label{fig:arch_ngram}
\end{figure}

Within each NSTB-MoSE block, window-based self-attention is adopted to limit attention cost while retaining non-local interactions within a window. Let $X \in \mathbb{R}^{H \times W \times C}$ denote an input feature map. Window partitioning yields
\begin{equation}
  X_{\text{win}} = \text{Partition}(X) \in \mathbb{R}^{N \times M^2 \times C}, \quad
  N = \frac{HW}{M^2},
  \label{eq:partition}
\end{equation}
where $M \times M$ is the window size. However, non-overlapping windows can introduce boundary discontinuities because tokens at the border of a window lack direct interaction with adjacent windows. To mitigate this limitation, N-Gram context is constructed by extracting overlapping local neighborhoods using a sliding operator:
\begin{equation}
  X_{\text{ngram}} = \text{Unfold}(X, \text{kernel}=n, \text{stride}=1).
  \label{eq:ngram}
\end{equation}
As illustrated in Fig.~\ref{fig:arch_ngram}, the context branch aggregates features from overlapping neighborhoods to provide each window token with adjacent information, promoting continuity in repetitive terrain textures.

The extracted context is projected to the window-token representation and fused with the window features prior to attention. This design injects local continuity cues without expanding the attention region, thereby improving boundary consistency while maintaining window-level efficiency:
\begin{equation}
  \tilde{X}_{\text{win}} = X_{\text{win}} + X_{\text{context}}(X_{\text{ngram}}),
  \label{eq:fusion}
\end{equation}
where $X_{\text{context}}(\cdot)$ denotes the context projection and aggregation function. Concretely, the overlapped features are first projected with pointwise ($1\times1$) convolutions to reduce channel dimensionality and constrain computational overhead. A lightweight sliding-window attention is then applied over $n \times n$ adjacent overlapping patches (with $n=2$ in this work) to aggregate inter-patch dependencies, followed by average pooling to form a localized context vector. The resulting context is added to the non-overlapping window tokens $X_{\text{win}}$ prior to self-attention, enriching each window with adjacent information and improving cross-window continuity. Self-attention with relative position bias and local positional encoding is then applied to $\tilde{X}_{\text{win}}$, followed by post-attention normalization to regularize the augmented representations:
\begin{equation}
  \hat{X} = X_{\text{in}} + \text{LayerNorm}\!\left(\text{Attention}(\tilde{X}_{\text{win}})\right).
  \label{eq:attn_norm}
\end{equation}

Transformer blocks typically use dense MLPs as the feed-forward pathway, which scales computation linearly with the hidden dimension and becomes a dominant cost when capacity is increased. To improve the capacity--efficiency trade-off, the dense MLP is replaced with a sparse Mixture-of-Experts (MoE) feed-forward network. The MoE layer comprises a gating function $G(\cdot)$ and a set of experts $\{E_i\}_{i=1}^{N_e}$. For each token $\hat{x}$, layer-normalized features are routed to the top-$K$ experts:
\begin{equation}
  p = G(\hat{x}_{\text{norm}}) = \text{Softmax}\!\left(\text{TopK}(\hat{x}_{\text{norm}} W_g, K)\right),
  \label{eq:gating}
\end{equation}
where $\hat{x}_{\text{norm}} = \text{LayerNorm}(\hat{x})$ and $W_g$ denotes the gating projection. Sparse routing activates only a subset of experts per token, enabling increased parameter capacity without proportional growth in inference cost. The MoE output is computed as a weighted mixture over the selected experts:
\begin{equation}
  y = \sum_{i \in \text{TopK}} p_i \, E_i(\hat{x}_{\text{norm}}).
  \label{eq:moe_out}
\end{equation}

After expert processing, token-wise routing may weaken local spatial coherence because neighboring tokens can be assigned to different experts and therefore receive different transformations. To regularize the spatial continuity of the MoE output, we apply a lightweight Smart Merger module after reassembling the token sequence into the feature-map layout. Specifically, the Smart Merger is implemented as a residual depth-wise spatial convolution. It is not a pointwise $1 \times 1$ convolution and does not use attention-based fusion. Instead, it performs channel-independent local spatial smoothing while preserving the sparse expert computation:
\begin{equation}
  y_{\text{merged}}
  =
  \operatorname{SmartMerger}(y)
  =
  \operatorname{DWConv}_{k \times k}(y) + y ,
  \label{eq:merger}
\end{equation}
where $k$ denotes the kernel size of the depth-wise convolution.
Finally, a residual connection yields the block output and supports stable optimization in deep networks:
\begin{equation}
  X_{\text{out}} = \hat{X} + y_{\text{merged}} .
  \label{eq:block_out}
\end{equation}

Computationally, the complexity of window-based multi-head self-attention (W-MSA) in a Swin-style block is
\begin{equation}
\Omega(\text{W-MSA}) = 4HWC^2 + 2M^2HWC,
\end{equation}
where $H \times W$ is the feature map resolution, $C$ is the channel dimension, and $M$ is the window size. The N-Gram context adds computation from context extraction and projection. For an N-Gram kernel size $n$, the additional complexity is approximately
\begin{equation}
\Omega(\text{N-Gram}) \approx 2n^2HWC,
\end{equation}
which remains moderate when $n$ is small relative to $M$. The MoE feed-forward pathway further decouples capacity from dense computation through sparse expert activation, improving the efficiency of scaling model capacity.

Training uses AdamW with $\beta_1 = 0.9$ and $\beta_2 = 0.999$. A weighted fusion objective supervises reconstruction quality and routing regularization:
\begin{equation}
\mathcal{L}_{\text{total}} =
\lambda_1 \mathcal{L}_{\text{pix}} +
\lambda_2 \mathcal{L}_{\text{ssim}} +
\lambda_3 \mathcal{L}_{\text{ncc}} +
\lambda_4 \mathcal{L}_{\text{perceptual}} +
\lambda_5 \mathcal{L}_{\text{moe}} .
\end{equation}
The loss terms are defined as follows. $\mathcal{L}_{\text{pix}}$ is an $\ell_1$ loss between $I_{SR}$ and the ground-truth image; $\mathcal{L}_{\text{ssim}}$ is an SSIM-based loss that penalizes local structural distortions; $\mathcal{L}_{\text{ncc}}$ is a normalized cross-correlation loss that encourages global statistical consistency; $\mathcal{L}_{\text{perceptual}}$ is an $\ell_1$ loss on deep features extracted from a fixed VGG-19 network up to \texttt{ReLU5\_4}; and $\mathcal{L}_{\text{moe}}$ is a router load-balancing regularizer that penalizes highly skewed expert usage and encourages balanced activation across experts. This objective balances pixel-level fidelity, structural similarity, perceptual consistency, and expert-routing regularization to support stable training and improved generalization.
\section{Experiments and Results}
\label{sec:experiments}

\begin{figure*}[h]
\centering
\small
\makebox[0.16\linewidth]{Full reference} \hfill
\makebox[0.16\linewidth]{LR Input} \hfill
\makebox[0.16\linewidth]{Bicubic} \hfill
\makebox[0.16\linewidth]{SwinIR} \hfill
\makebox[0.16\linewidth]{\textbf{NGram-MoSE}} \hfill
\makebox[0.16\linewidth]{Reference patch} \\
\vspace{2pt}

\includegraphics[width=0.16\linewidth]{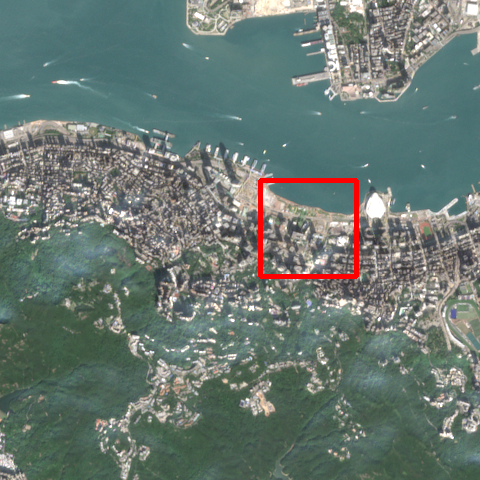} \hfill
\includegraphics[width=0.16\linewidth]{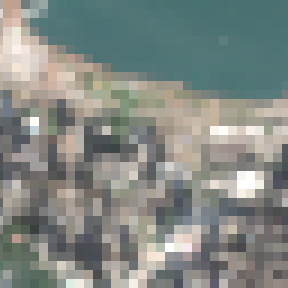} \hfill
\includegraphics[width=0.16\linewidth]{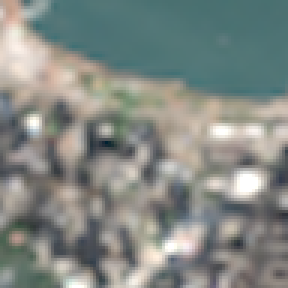} \hfill
\includegraphics[width=0.16\linewidth]{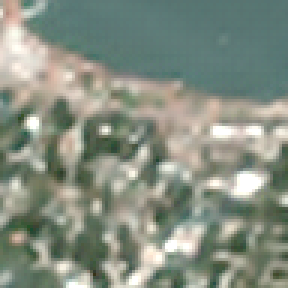} \hfill
\includegraphics[width=0.16\linewidth]{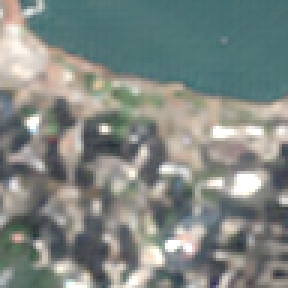} \hfill
\includegraphics[width=0.16\linewidth]{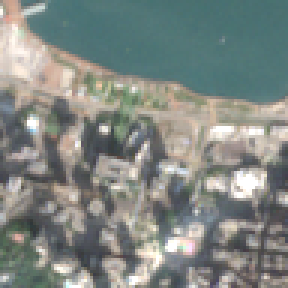} \\
\vspace{2pt}

\includegraphics[width=0.16\linewidth]{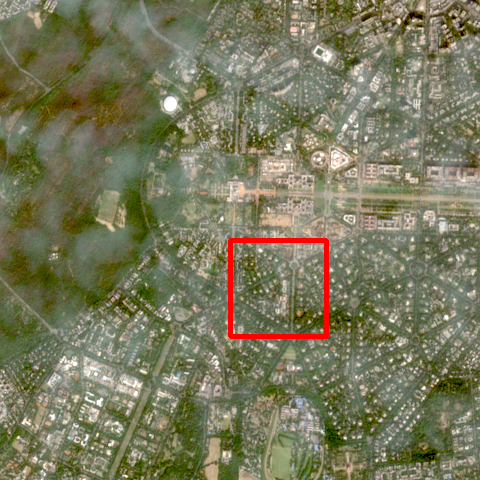} \hfill
\includegraphics[width=0.16\linewidth]{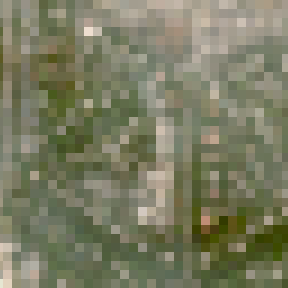} \hfill
\includegraphics[width=0.16\linewidth]{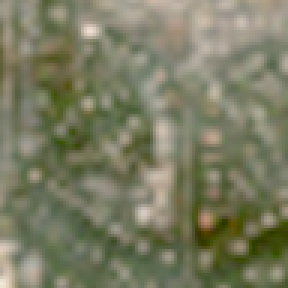} \hfill
\includegraphics[width=0.16\linewidth]{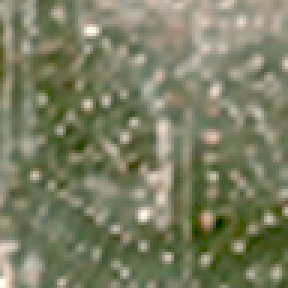} \hfill
\includegraphics[width=0.16\linewidth]{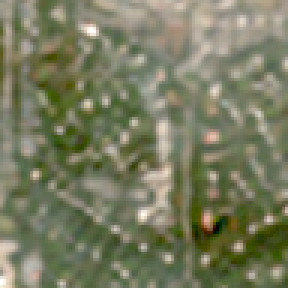} \hfill
\includegraphics[width=0.16\linewidth]{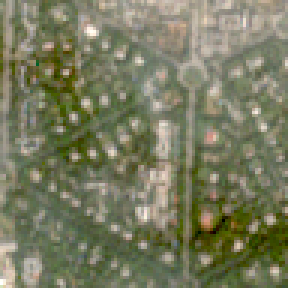} \\
\vspace{2pt}

\includegraphics[width=0.16\linewidth]{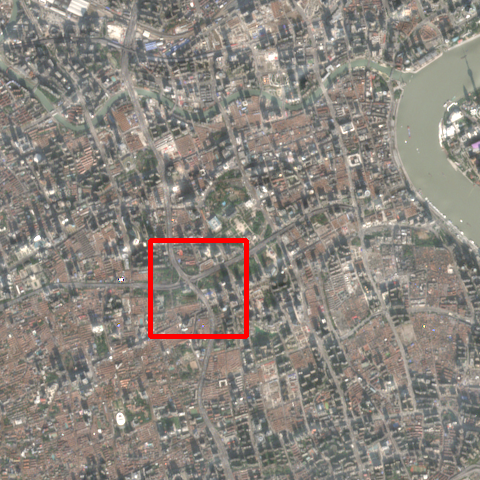} \hfill
\includegraphics[width=0.16\linewidth]{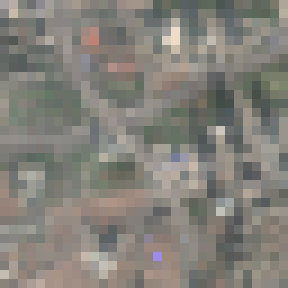} \hfill
\includegraphics[width=0.16\linewidth]{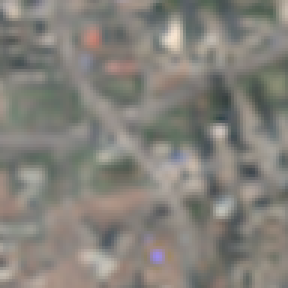} \hfill
\includegraphics[width=0.16\linewidth]{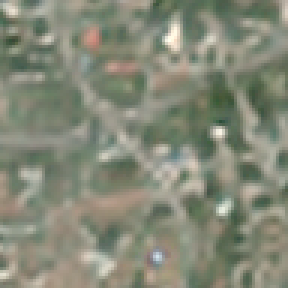} \hfill
\includegraphics[width=0.16\linewidth]{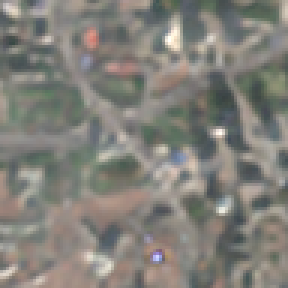} \hfill
\includegraphics[width=0.16\linewidth]{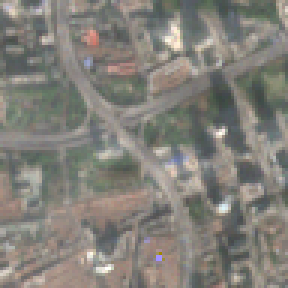} \\

\caption{\textbf{Qualitative results on representative urban OOD scenes ($3\times$ SR).} The leftmost column provides full-scene context with the zoomed region indicated. Columns compare degraded inputs and super-resolved outputs from different methods against the native-resolution reference.}
\label{fig:urban_ood_results}
\end{figure*}

This section reports the experimental protocol and quantitative restoration results, followed by application-driven evaluation on landslide detection and segmentation. The training set is constructed from five large-scale high-resolution remote sensing scenes ($5000 \times 5000$ pixels each) spanning urban and mountainous regions. Each scene is partitioned into a primary training pool of approximately 400 non-overlapping tiles of size $480 \times 480$. During optimization, random $128 \times 128$ sub-patches are sampled dynamically from these tiles to increase sample diversity and reduce spatial overfitting. Evaluation is conducted on an independent test set of 29 geographically disjoint scenes ($480 \times 480$), which are excluded from training to reduce spatial autocorrelation and provide an out-of-distribution (OOD) assessment.

Following standard SISR protocols, high-resolution images are downsampled by a factor of $\times 3$ to generate low-resolution inputs, and super-resolution is performed to reconstruct the corresponding high-resolution targets. Restoration quality is measured using PSNR and SSIM on the OOD test set. Computational complexity is reported in terms of parameter count and FLOPs, using SwinIR as a reference Transformer baseline. In addition to pixel-level metrics, semantic utility is evaluated via downstream landslide detection and segmentation using YOLOv8-Seg \cite{yolov8_ultralytics} on the Landslide4Sense benchmark \cite{Ghorbanzadeh_2022}. Downstream evaluation is conducted in two complementary settings: (i) restoration from degraded inputs back to the detector training scale, and (ii) scale extrapolation beyond the detector training scale to assess cross-scale robustness.

Table~\ref{tab:quantitative_psnr} summarizes restoration performance on the OOD test set. The results highlight the sensitivity of model scale to limited training diversity: the heavyweight SwinIR configuration used in this study (Depth=36, Dim=180) exhibits reduced generalization under geographically disjoint evaluation, yielding 31.31\,dB PSNR. In contrast, NGram-MoSE achieves 31.68\,dB PSNR and 0.9089 SSIM, indicating improved reconstruction fidelity under OOD terrains. Representative qualitative comparisons on urban OOD scenes are shown in Fig.~\ref{fig:urban_ood_results}, which complements the pixel-level metrics by illustrating typical reconstruction behaviors under domain shift.

\begin{table}[t]
\centering
\small
\caption{\textbf{Quantitative restoration performance on the OOD test set (scale $\times 3$).} Evaluation on 29 geographically disjoint scenes highlights generalization under limited training diversity.}
\label{tab:quantitative_psnr}
\begin{tabular}{l|c|cc}
\hline
\textbf{Method} & \textbf{Scale} & \textbf{PSNR (dB)} & \textbf{SSIM} \\
\hline
Bicubic & $\times 3$ & 31.43 & 0.8873 \\
SwinIR & $\times 3$ & 31.31 & 0.8946 \\
\textbf{NGram-MoSE} (Proposed) & $\times 3$ & \textbf{31.68} & \textbf{0.9089} \\
\hline
\end{tabular}
\end{table}

\begin{table}[t]
\centering
\caption{\textbf{Architectural complexity (Input: $128 \times 128$).} NGram-MoSE reduces FLOPs by 14$\times$ relative to the SwinIR reference.}
\label{tab:complexity}
\resizebox{\columnwidth}{!}{
\begin{tabular}{l|c|c|c}
\hline
\textbf{Method} & \textbf{Params (M)} & \textbf{FLOPs (G)} & \textbf{Reduction} \\ \hline
SwinIR & 11.89 & 195.34 & $1.0\times$ \\
\textbf{NGram-MoSE} (Proposed) & \textbf{1.70} & \textbf{13.94} & \textbf{14.0$\times$} \\
\hline
\end{tabular}
}
\end{table}

Model complexity is summarized in Table~\ref{tab:complexity}. NGram-MoSE requires 1.70\,M parameters and 13.94\,G FLOPs for a $128 \times 128$ input, compared to 11.89\,M parameters and 195.34\,G FLOPs for the SwinIR reference, corresponding to a 14$\times$ reduction in FLOPs. This efficiency supports deployment in resource-constrained remote sensing pipelines where compute and memory budgets are limited. The SwinIR model is reported as a heavyweight Transformer reference (Depth=36, Dim=180), whereas NGram-MoSE is configured as a lightweight architecture (Depth=14, Dim=64) to target efficiency-constrained settings.

\begin{figure*}[h]
\centering
\begin{subfigure}[b]{0.19\linewidth}
    \includegraphics[width=\linewidth]{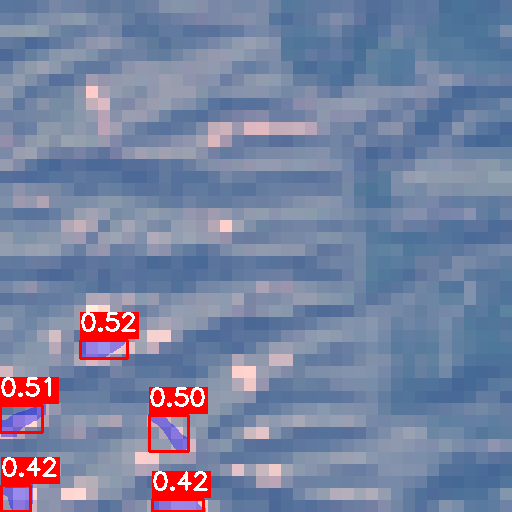}
    \caption{LR Input}
\end{subfigure}\hfill
\begin{subfigure}[b]{0.19\linewidth}
    \includegraphics[width=\linewidth]{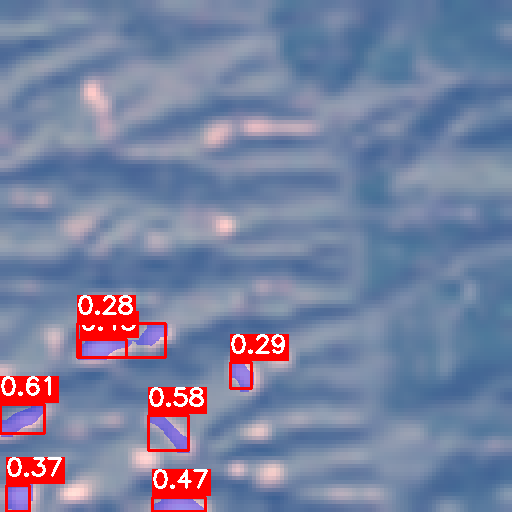}
    \caption{Bicubic}
\end{subfigure}\hfill
\begin{subfigure}[b]{0.19\linewidth}
    \includegraphics[width=\linewidth]{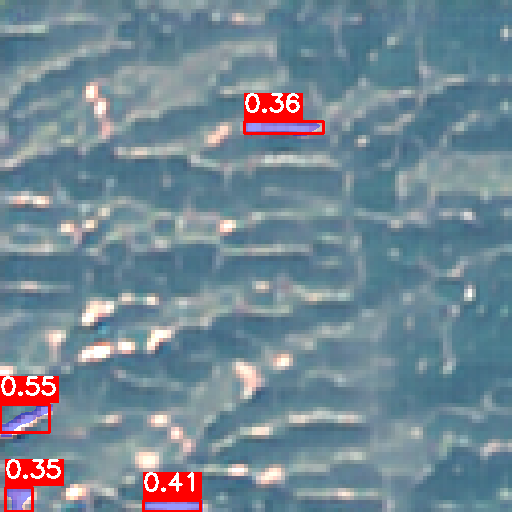}
    \caption{SwinIR}
\end{subfigure}\hfill
\begin{subfigure}[b]{0.19\linewidth}
    \includegraphics[width=\linewidth]{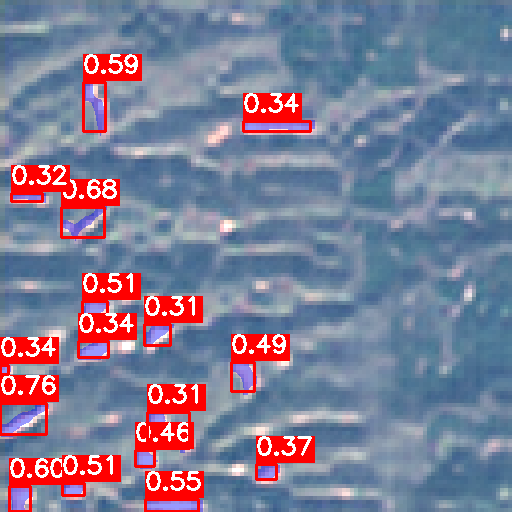}
    \caption{NGram-MoSE}
\end{subfigure}\hfill
\begin{subfigure}[b]{0.19\linewidth}
    \includegraphics[width=\linewidth]{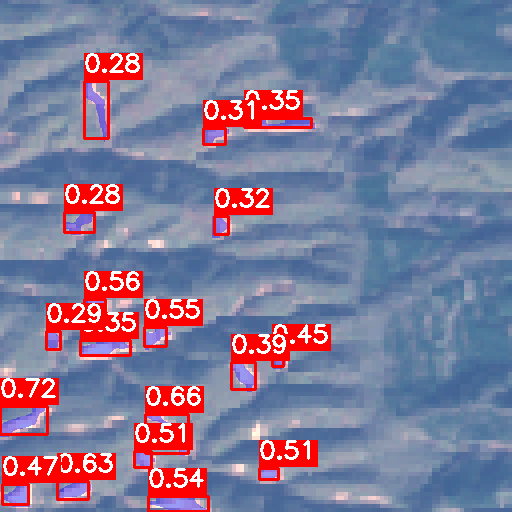}
    \caption{Reference}
\end{subfigure}

\caption{\textbf{Qualitative comparison for downstream evaluation at the detector training scale.} From left to right: degraded input, bicubic upsampling, SwinIR, NGram-MoSE, and the native-resolution reference.}
\label{fig:scenario_a_visual}
\end{figure*}

Downstream evaluation assesses whether restoration quality translates into improved semantic performance for landslide monitoring. The first setting evaluates restoration to the detector training scale. Native-resolution imagery is downsampled to simulate degraded inputs and restored back to the native scale using different SR methods. A YOLOv8-Seg model trained exclusively on native-resolution imagery is applied to the restored outputs. This protocol isolates whether reconstructed images remain semantically compatible with the original data distribution. Table~\ref{tab:scenario_a} shows that the degraded input yields 23.03\% mAP@50. Bicubic upsampling improves mAP@50 to 25.64\%, while SwinIR attains 25.61\%. NGram-MoSE increases mAP@50 to 30.11\%, corresponding to a 4.47\% absolute gain over Bicubic. Representative examples are shown in Fig.~\ref{fig:scenario_a_visual}.

\begin{table}[t]
\centering
\footnotesize
\caption{\textbf{Downstream evaluation at the detector training scale.} A detector trained on native-resolution imagery is evaluated on restored inputs generated from degraded observations.}
\label{tab:scenario_a}
\begin{tabular}{l|cc}
\hline
\textbf{Input Method} & \textbf{mAP@50} & \textbf{mAP@50-95} \\
\hline
Low-resolution input & 23.03\% & 8.01\% \\
Bicubic & 25.64\% & 9.49\% \\
SwinIR & 25.61\% & 8.47\% \\
\textbf{NGram-MoSE (Proposed)} & \textbf{30.11\%} & \textbf{10.49\%} \\
\hline
Native-resolution reference & 43.88\% & 19.26\% \\
\hline
\end{tabular}
\end{table}

\begin{table}[t]
\centering
\footnotesize
\caption{\textbf{Cross-scale robustness under scale extrapolation.} B3 evaluates performance in the upsampled domain, while B1 evaluates each detector on native-resolution inputs as a consistency check.}
\label{tab:scenario_b}
\setlength{\tabcolsep}{6pt}
\begin{tabular}{l|c|c}
\hline
\textbf{SwinIR-trained system} & \textbf{mAP@50} & \textbf{mAP@50-95} \\ \hline
B1: Native-resolution input & 34.23\% & 13.80\% \\
B2: Bicubic-upsampled input & 33.65\% & 13.52\% \\
B3: SwinIR-upsampled input & 42.79\% & 18.50\% \\ \hline
\multicolumn{3}{c}{} \\ [-0.5em]
\hline
\textbf{NGram-MoSE-trained system} & \textbf{mAP@50} & \textbf{mAP@50-95} \\ \hline
B1: Native-resolution input & \textbf{39.65\%} & \textbf{16.38\%} \\
B2: Bicubic-upsampled input & \textbf{38.66\%} & \textbf{16.04\%} \\
B3: NGram-MoSE-upsampled input & \textbf{43.41\%} & \textbf{18.34\%} \\ \hline
\end{tabular}
\end{table}

\begin{figure*}[hbt]
\centering
\begin{minipage}[t]{0.7\textwidth}
\makebox[0.24\linewidth]{\textbf{Input}} \hfill
\makebox[0.24\linewidth]{\textbf{Bicubic}} \hfill
\makebox[0.24\linewidth]{\textbf{SwinIR}} \hfill
\makebox[0.24\linewidth]{\textbf{NGram-MoSE}} \\

\includegraphics[width=0.24\linewidth]{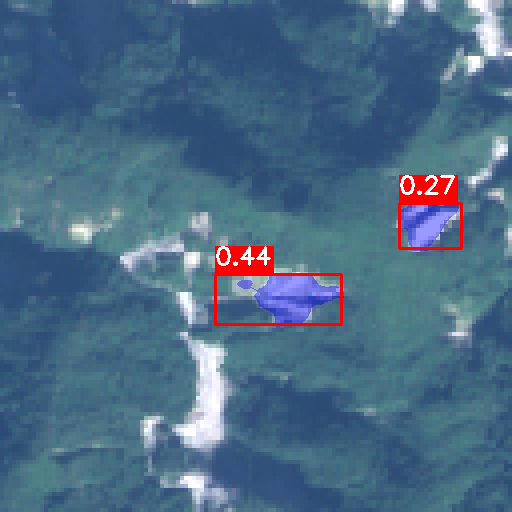}\hfill
\includegraphics[width=0.24\linewidth]{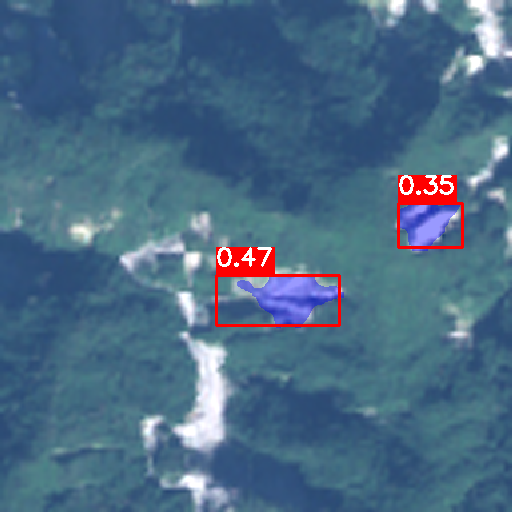}\hfill
\includegraphics[width=0.24\linewidth]{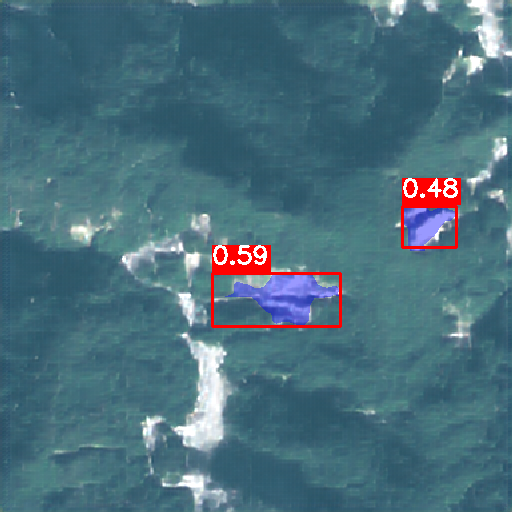}\hfill
\includegraphics[width=0.24\linewidth]{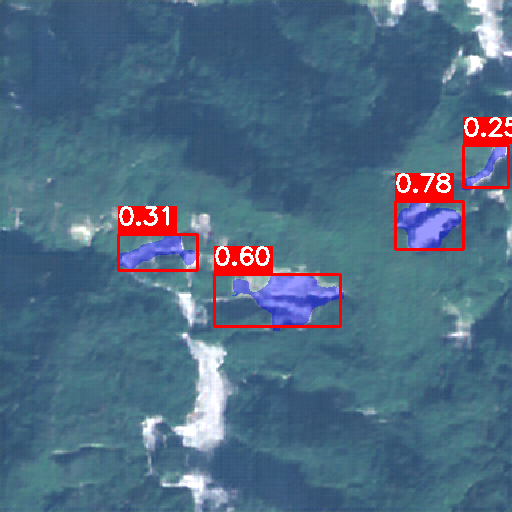}\\

\includegraphics[width=0.24\linewidth]{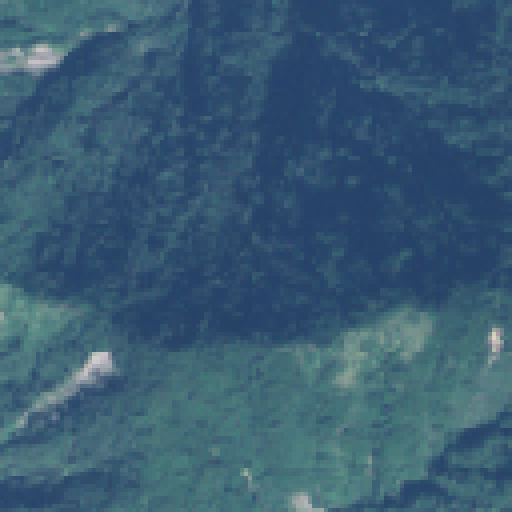}\hfill
\includegraphics[width=0.24\linewidth]{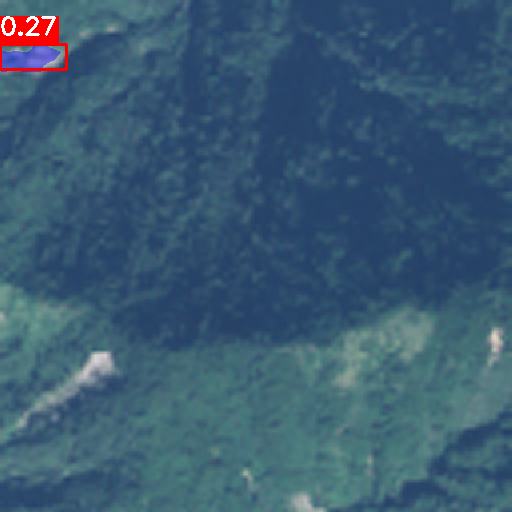}\hfill
\includegraphics[width=0.24\linewidth]{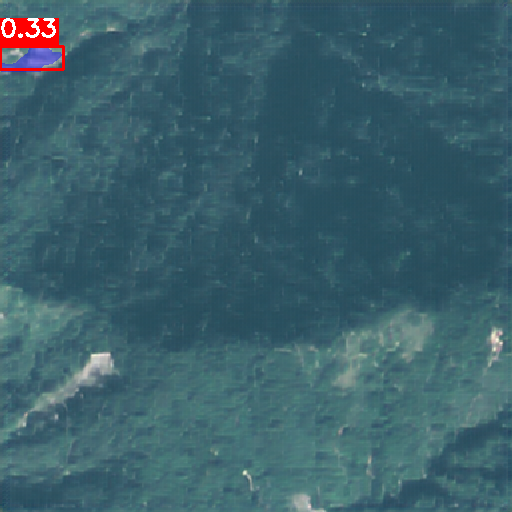}\hfill
\includegraphics[width=0.24\linewidth]{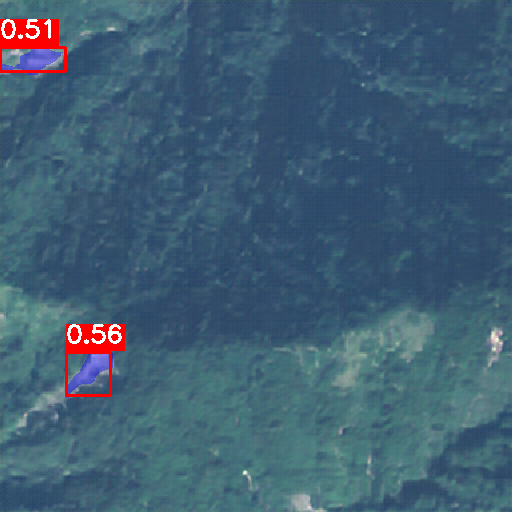}\\
\end{minipage}

\caption{\textbf{Qualitative results for cross-scale robustness under scale extrapolation.} Columns compare native-resolution inputs and $3\times$ upsampled outputs generated by different methods. The examples highlight representative behavior under cross-scale extrapolation.}
\label{fig:scenario_b_visual}
\end{figure*}

The second setting evaluates cross-scale robustness under scale extrapolation, where inputs are upsampled beyond the native resolution and corresponding higher-resolution ground truth is unavailable. Two separate detection systems are constructed to reduce train--test mismatch: one YOLOv8-Seg detector is trained on images upsampled by SwinIR and another detector is trained on images upsampled by NGram-MoSE. Each detector is evaluated in the corresponding upsampled domain (B3), where performance may reflect adaptation to SR-specific statistics. To quantify semantic drift, each detector is additionally evaluated on native-resolution inputs (B1) as a consistency check for compatibility with real imagery. Table~\ref{tab:scenario_b} shows that the SwinIR-trained system exhibits a larger decrease from B3 (42.79\%) to B1 (34.23\%), whereas the NGram-MoSE-trained system maintains stronger B1 performance (39.65\%). Representative qualitative comparisons are shown in Fig.~\ref{fig:scenario_b_visual}.
\section{Conclusion}
\label{sec:conclusion}

This work presents NGram-MoSE, a lightweight Transformer architecture for remote sensing image super-resolution (SR) that balances restoration fidelity with computational efficiency. The design addresses two practical challenges in window-based Transformer SR: boundary discontinuities that disrupt continuous terrain textures and the high cost of dense feed-forward computation when model capacity is increased. To mitigate these issues, NGram-MoSE integrates N-Gram Context Injection to enhance cross-window local continuity and a sparse Mixture-of-Experts (MoE) mechanism to scale capacity through conditional computation.

Experiments on a geographically disjoint out-of-distribution (OOD) test set show that NGram-MoSE improves restoration fidelity while substantially reducing computation relative to SwinIR. Under the reported configurations, NGram-MoSE achieves 31.68\,dB PSNR and reduces FLOPs by \(14\times\) compared to SwinIR. Beyond pixel-level metrics, downstream evaluation on the Landslide4Sense benchmark indicates that restored imagery can improve landslide detection and segmentation performance: restoring degraded inputs to the detector training scale yields a 4.47\% absolute gain in mAP@50 over bicubic upsampling. In addition, scale extrapolation experiments show stronger cross-scale consistency under native-resolution consistency checks, suggesting improved robustness to scale-induced distribution shift.

Future work may incorporate higher-resolution imagery to provide stronger supervision for explicit sub-native SR and to further evaluate cross-sensor generalization. Overall, the results suggest that combining local continuity modeling with sparse capacity scaling offers an effective and practical SR solution for remote sensing pipelines operating under limited compute budgets and constrained training diversity.

\bibliographystyle{IEEEtran}
\bibliography{main}

\end{document}